\pdfoutput=1

\documentclass[11pt]{article}

\usepackage[]{acl}

\usepackage{times}
\usepackage{graphicx} 
\usepackage{latexsym}
\usepackage{float} 
\usepackage{amsmath}
\usepackage{enumitem}
\usepackage{tabu}

\usepackage{booktabs}
\usepackage{tabularx}
\usepackage{soul}

\usepackage[T1]{fontenc}

\usepackage[utf8]{inputenc}

\usepackage{microtype}

\title{ A Simple Approach to Jointly Rank Passages and Select Relevant Sentences in the OBQA Context}

\author{Man Luo\\
  Arizona State University \\ 
  \texttt{mluo26@asu.edu} \\\And
  Shuguang Chen \\
  University of Houston \\
  \texttt{schen52@uh.edu} \\\And
  Chitta Baral\\
  Arizona State University \\ 
  \texttt{chitta@asu.edu} }

\begin{document}
\maketitle
\begin{abstract}
In the open book question answering (OBQA) task, selecting the relevant passages and sentences from distracting information is crucial to reason the answer to a question. HotpotQA dataset is designed to teach and evaluate systems to do both passage ranking and sentence selection. Many existing frameworks use separate models to select relevant passages and sentences respectively.
Such systems not only have high complexity in terms of the parameters of models but also fail to take the advantage of training these two tasks together since one task can be beneficial for the other one. 
In this work, we present a simple yet effective framework to address these limitations by jointly ranking passages and selecting sentences. 
Furthermore, we propose consistency and similarity constraints to promote the correlation and interaction between passage ranking and sentence selection.The experiments demonstrate that our framework can achieve competitive results with previous systems and outperform the baseline by 28\% in terms of exact matching of relevant sentences on the HotpotQA dataset.

\end{abstract}

\section{Introduction}

Open book question answering (OBQA) requires a system to find the relevant documents to reason the answer to a question.
It has wide and practical Natural Language Processing (NLP) applications such as search engines~\citep{kwiatkowski-etal-2019-natural} and dialogue systems~\citep{reddy-etal-2019-coqa,choi-etal-2018-quac}. 
Among several OBQA datasets~\citep{Dhingra2017QuasarDF,Mihaylov2018CanAS,Khot2020QASCAD},   HotpotQA~\cite{yang2018HotpotQA} is more challenging because it requires a system not only to find the relevant passages from large corpus but also find the relevant sentences in the passage which eventually reach to the answer. Such a task also increases the interpretability of the systems.  


\begin{figure}[t]
    \centering
    \includegraphics[width=1\linewidth]{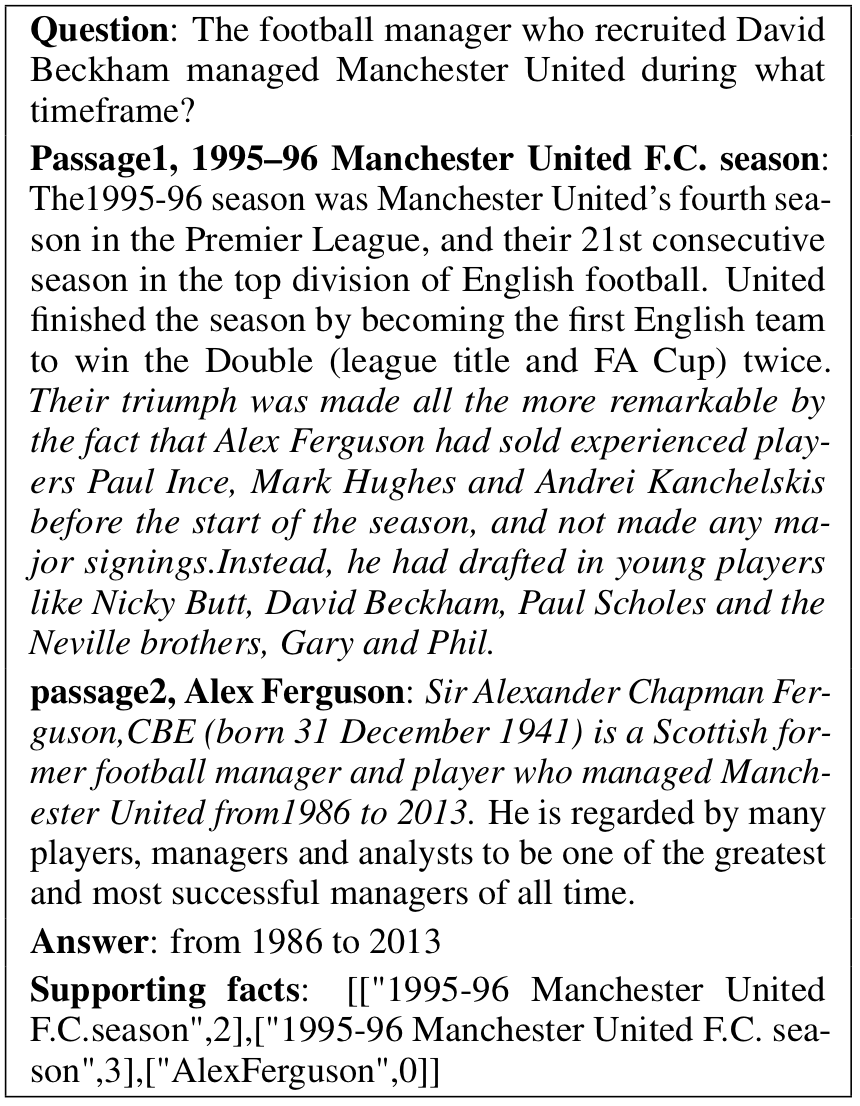}
    \caption{An example from the HotpotQA dataset, where the question should be answered by combining supporting facts(SP) from two passages. In the SP, the first string refers to the title of passage, and the second integer means the index of the sentence.}
    \label{fig: HotpotQA}
\end{figure}

To address this challenge, most of the previous work~\citep{nie2019revealing,fang2019hierarchical,Tu2020SelectAA,groeneveld2020simple} use two-step pipeline: identify the most relevant passage by one model and then match each question with a single sentence in the corresponding passage by another model. 
Such systems are heavy in terms of the size of the models which requires long training and inference time. 
Green AI has recently been advocated to against the trend of building large models which are both environmentally unfriendly and expensive, raising
barriers to participation in NLP research~\cite{schwartz2020green}.
Apparently, systems using multiple models to solve HotpotQA task do not belong to the family of Green AI.
Furthermore, the benefits of learning from passage ranking and selecting relevant sentences are not well utilized by these systems. 
Intuitively, if a passage is ranked high, then some sentences in the passage should be selected as relevant. 
On the other hand, if a passage is ranked low, then all sentences in the passage should be classified as irrelevant.

To build a Green AI system and take advantage of multi-task learning, we introduce a Two-in-One model, a simple model trained on passage ranking and sentence selection jointly. 
More specifically, our model generates passage representations and sentence representations simultaneously, which are then fed to a passage ranker and sentence classifier respectively. 
Then we promote the interaction between passage ranking and sentence classification using consistency and similarity constraints. 
The consistency constraint is to enforce that the relevant passage includes relevant sentences, while the similarity constraint ensures the model to generate the representation of relevant passages more closer to the representations for relevant sentences than irrelevant ones. 
The experiments conducted on the HotpotQA datasets demonstrate that our simple model achieves competitive results with previous systems and outperforms the baselines by 28\%.



\section{Related Work}

\paragraph{HotpotQA Systems} A straightforward way to solve the HotpotQA challenge is to build a hierarchical system~\cite{nie2019revealing}, meaning a system first ranks relevant passages and then identifies relevant sentences from the selected passages. Such a hierarchical system involves multiple models thus requires long inference time. More importantly, such a system only leverages the impact of passage ranking on sentence selection but ignores the influence of the sentence selection on the passage ranking. Our framework achieves these two tasks by one model and facilitates the interaction by two constraints. \citet{groeneveld2020simple} proposes a pipeline based on three BERT models~\cite{devlin-etal-2019-bert} to solve the HotpotQA challenge. The system first selects relevant sentences and then detects the answer span, finally, identifies the relevant sentences according to the answer span. Though the pipeline is strong, the way it solves the problem is opposite to human beings. We, humans, identify the relevant sentences, and then give the answer span. Many existing works demonstrate the effectiveness of graph neural networks(GNN) on HotpotQA challenge~\citep{fang2019hierarchical,Tu2020SelectAA}. Since GNN is out of the scope of this work, we do not compare it with these frameworks.

\paragraph{Joint Model for QA}
Joint learning has been studied in Question Answering Tasks. \citet{Deng2020JointLO} proposes a joint model to tackle community question answering such that the model can simultaneously select the set of correct answers from candidates and generate an abstractive summary for each selected answer. \citet{sun2019joint} proposes a generative collaborative network to answer questions and generate questions. The main difference between our work and previous ones  are in two sense (1) our proposed model uses the shared encoder to tackle two classification tasks (2) besides the loss function to optimize individual tasks, we also propose two constraints that utilize the relation between these two tasks.


\section{HotpotQA Dataset}\label{sec:hotpotqa}

HotpotQA dataset~\citep{yang2018HotpotQA} is designed for multi-hop reasoning question answering tasks, i.e. to reason over multiple documents and answer questions (see Figure \ref{fig: HotpotQA}). Particularly, HotpotQA challenge requires reasoning over two passages. Furthermore, to guide the system to perform meaningful and explainable reasoning, the dataset also provides supporting facts (SP) that reach the answer to the question. 
HotpotQA provide two challenging settings: in \textbf{Fullwiki setting}, a system needs to rank passage from the entire wiki corpus; in \textbf{Distractor setting}, 10 distracting passages (including relevant ones) are given for each question. In this work, we mainly focus on the latter setting. From the training set, we find that 70.4\% questions have exactly two supporting facts (SP), and 60.0\% of SP are the first sentence of passages.

\section{Method}

We aim to jointly conduct two tasks, passage ranking  and supporting facts selection for HotpotQA. Given a question Q, the goal is to simultaneously rank the set of 
candidates A = \{$a_1$, ..., $a_i$\} and identify the supporting facts for the TopK\footnote{The value of K depends on the task, and for HotpotQA, K is 2.} passages.
\subsection{Model: Two-in-One Framework}
\begin{figure}[h]
    \centering
    \includegraphics[width=0.9\linewidth]{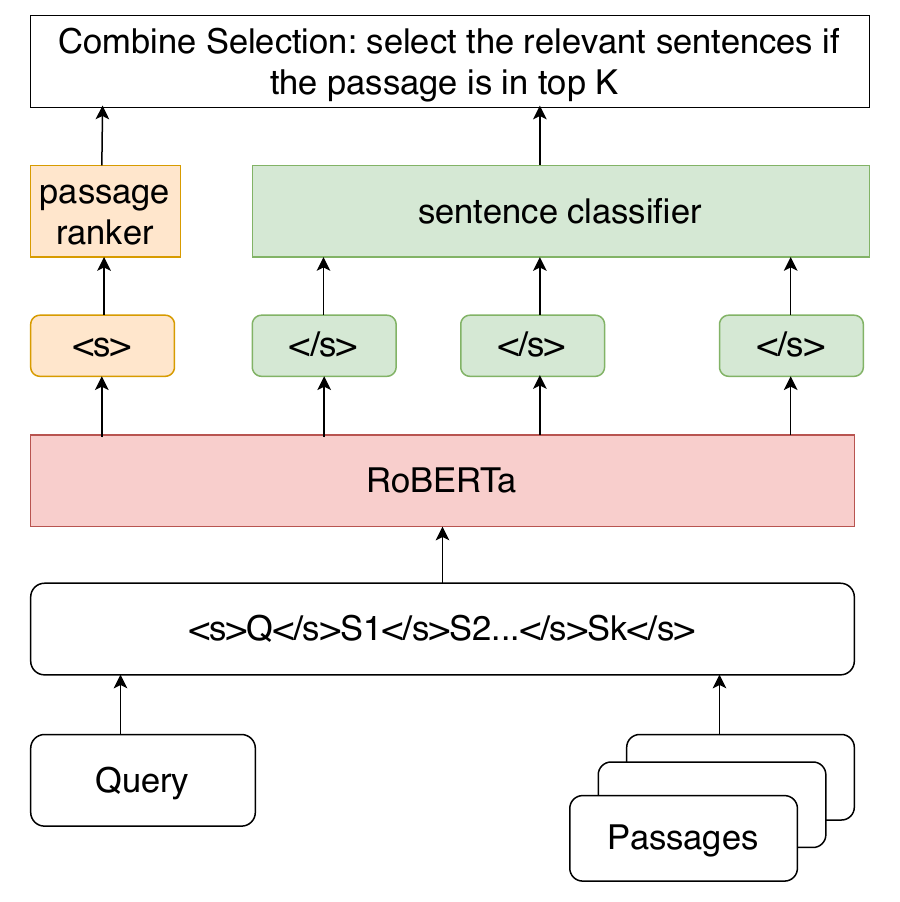}
    \caption{The architecture of Two-in-One model for passage ranking and relevant sentence selection. For HotpotQA dataset, K is two.}
    \label{fig:architecture}
\end{figure}

We introduce the proposed joint model for passage ranking and support fact selection, Two-in-One, which uses state-of-the-art transformer-based model~\citep{NIPS2017_3f5ee243} to encode questions and contexts. In this work, we use RoBERTa~\citep{liu2019roberta}, however, any other variants like ELECTRA~\citep{clark2020electra} can be applied in this framework. 
The model architecture is given in Figure \ref{fig:architecture}. On top of the encoder, there are two MLP layers to score passages and sentences respectively. In details, given a question and a passage, we firstly create an input to feed through  RoBERTa \cite{liu2019roberta}  by concatenating the question and the passage as follows, 
$\langle s \rangle Q \langle /s \rangle S_{1}\langle /s\rangle S_{2}...\langle /s\rangle S_{k} \langle /s \rangle $
where $ \langle s \rangle$ and  $ \langle /s \rangle$ are special tokens in RoBERTa,  $S_i$ is the $i^{th}$ sentence from a passage. 
We take $ \langle s \rangle$ as the contextual representation for passage ranking and the $ \langle /s \rangle$ in front of each sentence for sentence selection. The passage ranker and the sentence classifier have identical structure (two-layer Multiple-Layer Perceptron(MLP)) but different weights.


 \label{apd:achitecture}

The model is jointly trained by passage loss and sentence loss. In detail, during the training time, we assign the relevant passages and sentences with ground truth score 1 while irrelevant passages and sentences with ground truth score -1. Then, Mean Square Error(MSE) loss is applied to calculate the passage and sentence loss as follows,
\begin{equation}
    \begin{split}
        &\mathcal{L}^{pass} = (\hat{y}-y)^2 ,\\
        &\mathcal{L}^{sent} = \sum_{i=1}^{K}(\hat{x_{i}}-x_i)^2 ,\\
        &\mathcal{L}^{joint} = \mathcal{L}^{pass}+\mathcal{L}^{sent} ,
    \end{split}
\label{eq:joint}
\end{equation}
where $\hat{y}$ is the predicted passage score, $y$ is the ground truth score of the passage, $\hat{x_i}$ and $x_i$ are the predicted sentence score and ground truth score of $S_i$, respectively, and $K$ is the total number of sentences in the passage. We simply sum up the passage loss and sentence loss to jointly update model parameters.

During the inference time, passages are ranked based on the logits given by the passage ranker. For the sentence classification, we take 0\footnote{The reason for threshold ``0'' is that it is the middle value of 1 and -1, which are labels for relevant and irrelevant sentences in the training time.} as the threshold to classify the relevance of each sentence: if the score given by the sentence classifier is larger than 0, then it is relevant; otherwise, irrelevant.


Next, we introduce two constraints to facilitate the interaction between these two tasks.

\subsection {Consistency Constraint}
Intuitively, if a passage is relevant to the question, then there are some sentences from the passages that are relevant; on the other hand, if a passage is not relevant to the answer, then there should not be relevant sentences inside the passage. Thus, we propose a consistency constraint over the passage ranker and sentence classifier to minimize the gap between the passage score and the maximum sentence score. The loss function is as follows:
\begin{equation}
\mathcal{L}^{con} = (\hat{y} - max(\mathbf{x}))^2,
\end{equation}
where $\mathbf{x} = [\hat{x}_1 \dots \hat{x}_n]$ denotes a stack of predicted sentence scores.

\subsection {Similarity Constraint}
As we have shown at the beginning of this section,  token $\langle s \rangle$ is used to get the passage score, and each  token $\langle /s \rangle$ is used to get the sentence score. Intuitively, the similarity between token $\langle s \rangle$ of a relevant passage is more close to token $\langle /s \rangle$ of a relevant sentence than to  $\langle /s \rangle$ of any irrelevant sentence. To enforce this constraint, we use triplet as follows:
\begin{equation}
\label{eq:sim_constraint}
    \begin{split}
       \mathcal{L}^{sim} = \frac{1}{N\cdot M}\sum_{i=1}^{N} \sum_{j=1}^{M} (&max\{d(v^{p}, v^{r}_{i}) \\ &- d(v^{p}, v^{n}_{j})+m, 0 \}), 
    \end{split}
\end{equation}
where $d(\cdot, \cdot)$ is the Euclidian similarity, $N$ is the number of relevant sentences,  $M$ is the number of irrelevant sentences,  $v^p,  v^r, v^n$ is the vector representation of the relevant passage, relevant sentence, and irrelevant sentence respectively. 
Equation \ref{eq:sim_constraint} enforces that all the relevant sentences should have higher similarity with the passage than all the irrelevant sentences by a margin $m$; otherwise, the model would be penalized.
In practice, we set the margin $m$ at 1 and find optimum results. We train our model in an end-to-end fashion by combining
$\mathcal{L}^{joint}$, $\mathcal{L}^{con}$ and $\mathcal{L}^{dis}$.

\begin{table*}[t]
\centering
\resizebox{\textwidth}{!}{
\begin{tabular}{ccccccc}
\hline
\textbf{Model} & \textbf{\# Parameters} &\textbf{SP Precision } & \textbf{SP Recall }  & \textbf{SP F1 } &   \textbf{SP EM } & \textbf{Passage EM}\\
\hline
Sentence Selection Baseline & $\sim$330M & 67.96 & 81.05 & 72.02 & 28.12 & 69.70 \\
Passage Selection Baseline & $\sim$330M & 66.43 & 56.55 & 60.20 & 27.30 & 90.44 \\
Two-in-One + sim (Ours) & $\sim$330M & \textbf{88.06} & \textbf{85.68 }& \textbf{85.82} & \textbf{59.17} & \textbf{91.11} \\
\midrule
QUARK  & $\sim$1020M$^{*}$ & N/A & N/A & 86.97& 60.72 & N/A \\
SAE(RoBERTa) & $\sim$660M+$^{*}$ & N/A & N/A & 87.38& \textbf{63.30} & N/A\\
HGN(RoBERTa) & $\sim$330M+$^{*}$ & N/A & N/A & \textbf{87.93} & N/A & N/A\\
\hline
\end{tabular}
}
\caption{\label{exp-result}
The Results for two baselines and Two-in-One model with similarity constraint on dev set of HotpotQA distracting dataset. SP stands for supporting facts and EM for Exact Match. $^{*}$ refers to estimation.
The bottom systems have much larger model size than our method, where QUARK~\cite{groeneveld2020simple}, is the result of a framework with 3 BERT models, SAE ~\cite{Tu2020SelectAA} uses two large language models and an GNN model, and HGN~\cite{fang2019hierarchical} uses a large language model, a GNN model and other reasoning layers. 
}
\end{table*}

\section{Experiment}

In this section, we first describe the training setup, and then introduce two baselines. We evaluate the two baselines and our proposed joint model on the HotpotQA dataset. \citet{yang2018HotpotQA} provides two metrics for supporting facts evaluation, exact matching (EM) and F1 score. We also present the precision and recall of SP, and the exact matching of passages for detailed comparison. We mainly compare our model with the QUARK system~\cite{groeneveld2020simple} since both QUARK and our method simply use language models without involving complicated reasoning models. For reference, we also present other state-of-the-art models in Table \ref{exp-result}.
Lastly, we conduct an ablation study to show the effectiveness of the proposed similarity loss and consistent loss.

\subsection{Experiment Setup}
We use Huggingface~\citep{wolf-etal-2020-transformers} and Pytorch~\citep{NEURIPS2019_9015} libraries to implement each model.
We use 4 TX1080 and V100 NVIDIA to train models in 5 epochs with a learning rate of 1e-5, batch size of 32.
We set the maximum input length in training to be 512. 

\subsection{Baseline}
To have comparable size of the model, two baselines have similar structure as our Two-in-One model. Our model has two classification heads, whereas each of the baselines has one classification head. One baseline is to select relevant sentences, and the other one is to rank passages. 

\paragraph{Sentence Selection Baseline } The first baseline is to select relevant sentence, and particularly, we use a RoBERTa-large with an additional MLP trained on question and a single sentence: $\langle s \rangle Q \langle /s \rangle S  \langle /s \rangle$,  where  $Q$ is a question and $S$ is a sentence. Although this model can not predict the relevant passage directly, based on the assumption that relevant passages include relevant sentences, we pick up two relevant passages based on the top2 sentence scores. When the top1 and the top2 sentences are from the same passage, we continue searching based on the ranking sentence scores until the second document comes up. Then the supporting facts are those sentences from the relevant documents with a score larger than 0.

\paragraph{Passage Selection Baseline} In the second baseline, again, we use RoBERTa-large but with the goal of passage selection. The input to the model is a question and a passage: $\langle s \rangle Q \langle /s \rangle P  \langle /s \rangle$. Since such a model can not predict sentence relevancy score, based on the statistic of HotpotQA that majority of training set has two supporting facts and the most of them are the first  sentences in a paragraph (see Section \ref{sec:hotpotqa}), we select supporting facts by the first sentence of the top1 and top2 passages.

\subsection{Result}

As we see from Table \ref{exp-result}, Two-in-One framework outperforms two baselines with large-margin improvement in all metrics, especially we see a significant improvement on the EM of SP. Our framework outperforms the Sentence Selection Baseline by 20\% and 4.5\% improvement on the precision and recall of SP, respectively, which demonstrates that jointly learning is beneficial for sentence classification. Also, jointly learning benefits for the passage ranking by comparing Two-in-One with Passage Selection Baseline on the EM of passage. Besides, we also compare Two-in-One with QUARK~\cite{groeneveld2020simple}, a framework involving three BERT models, (roughly three times larger than ours). Two-in-One achieves comparable results in terms of F1 and EM of SP regardless of much less parameters in our system. Notice that we do not have the other three values because they are not presented in their original paper.

\subsection{Ablation}
To evaluate the impacts of the consistency constraint and the similarity constraint, we conduct experiments with and without constraints. From Table \ref{tab:ablation}, we see that both consistency constraint and similarity constraint improve F1 and EM of SP and the similarity constraint also improves the EM of passages. We found that without any constraint, though the model can rank the passages well, it suffers from distinguishing between close sentences. The similarity constraint addresses this issue in some sense by maximizing the distance between relevant and irrelevant sentences.

\begin{table}[t]
\centering
\resizebox{\linewidth}{!} {
\begin{tabular}{lccc}
\hline
\textbf{Model} & \textbf{SP F1} & \textbf{SP EM} & \textbf{Passage EM}\\
\hline
Two-in-One  & 85.52 & 58.67  & 90.93 \\
Two-in-One + con & 85.55 &58.98  & 90.29 \\
Two-in-One + sim & {\bf 85.82} & {\bf 59.17}  & {\bf 91.11}\\
Two-in-One + con + sim  & 85.63 & 58.74 & 90.78 \\\hline
\end{tabular}
}
\caption{The results for Two-in-One model with or without consistency and similarity constraints.}
\label{tab:ablation}
\end{table}

To better understand the impact of consistency constraint, we analyze the consistency between the passage score and the sentence score.
The prediction of a model is consistent if the passage score agrees with the sentence scores and the agreement can be measured by the gap between the passage score and the maximum sentence score among all sentences in that passage.
We observe that by adding the consistency constraint, the gap between the passage score and the sentence score is much smaller than without the consistency constraint, i.e. 0.03 v.s. 0.11.  It demonstrates that the constraint is beneficial for consistent prediction.

\section{Future Work}

\paragraph{Model Architecture} It is easy to extend the Two-in-One model to Three-in-One model such that besides the passage ranking and sentence selection modules, a third module can predict the answer span. 
Like the simple extractive QA model based on RoBERTa, where a linear layer or an MLP can predict the start and end position of the answer span. 
A restricted inference procedure can be enforced that the answer span should be predicted from the selected sentence given by the previous model. 
One benefit is to reduce the difficulty for the answer selection model since less sentences will be seen by the model and the second benefit is to increase the interpretability of the model. 
On the other hand, if the sentence selection model makes mistakes, then such errors will carry to the answer span model which yields the wrong answer eventually.  



\paragraph{Apply to Full Open Domain Setting}
We only study the distracting setting of HotpotQA in this work, where 10 passages are already given for each question. 
On the other hand, in the full open domain setting, the passages need to be chosen from a large corpus. 
A simple approach to adapt Two-in-One model to the later setting is to use a retriever~\cite{robertson2009probabilistic,karpukhin2020dense,luo2022improving} to select the 10 passages and ask Two-in-one to choose the right passages and supporting facts.

\paragraph{Zero-shot Testing} 
Generalization is one of the major research concerns for neural networks~\cite{gokhale2022generalized}.
It is interesting to see how well the proposed two-in-One model performs in unseen domains.

\section{Conclusion}

In this work, we present a simple model, Two-in-One, to rank passage and classify sentence together.
By jointly training with passage ranking and sentence selection, the model is capable of capturing the correlation between passages and sentences. We show the effectiveness of our proposed framework by evaluating the model performance on the HotpotQA datasets, concluding that jointly modeling passage ranking and sentence selection is beneficial for the task of OBQA. Compared to the existing QA systems, our model,  with fewer parameters and more green than previous models, can achieve competitive results. 
We also propose multiple future directions to improve our model such as exploring the  relationship among passages, supporting sentences, and answers in modeling and generalizing our method on more datasets.

\bibliography{anthology,custom}


\end{document}